\documentclass{article}
\usepackage{times}
\usepackage{graphicx}
\usepackage{subfigure}
\usepackage{algorithm}
\usepackage{algorithmic}
\usepackage[unicode=true]{hyperref}

\usepackage{amsmath}
\usepackage{url}

\title{Efficient batchwise dropout training using submatrices}
\author{
  Ben Graham\\
  \texttt{b.graham@warwick.ac.uk}
  \and
  Jeremy Reizenstein\\
  \texttt{j.f.reizenstein@warwick.ac.uk}
  \and
  Leigh Robinson\\
  \texttt{leigh.robinson@warwick.ac.uk}
}
\begin{document}
\maketitle

\begin{abstract}
\noindent Dropout is a popular technique for regularizing artificial neural
networks. Dropout networks are generally trained by minibatch gradient
descent with a dropout mask turning off some of the units---a different
pattern of dropout is applied to every sample in the minibatch. We
explore a very simple alternative to the dropout mask. Instead of
masking dropped out units by setting them to zero, we perform matrix
multiplication using a submatrix of the weight matrix---unneeded hidden
units are never calculated. Performing dropout \emph{batchwise}, so
that one pattern of dropout is used for each sample in a minibatch,
we can substantially reduce training times. Batchwise dropout can
be used with fully-connected and convolutional neural networks.
\end{abstract}

\section{Independent versus batchwise dropout}

Dropout is a technique to regularize artificial neural networks---it
prevents overfitting \cite{dropout}. A fully connected network with
two hidden layers of 80 units each can learn to classify the MNIST
training set perfectly in about 20 training epochs---unfortunately
the test error is quite high, about 2\%. Increasing the number of
hidden units by a factor of 10 and using dropout results in a lower
test error, about 1.1\%. The dropout network takes longer to train
in two senses: each training epoch takes several times longer, and
the number of training epochs needed increases too. We consider a
technique for speeding up training with dropout---it can substantially
reduce the time needed per epoch.

Consider a very simple $\ell$-layer fully connected neural network
with dropout. To train it with a minibatch of $b$ samples, the forward
pass is described by the equations:
\[
x_{k+1}=[x_{k}\cdot d_{k}]\times W_{k}\qquad k=0,\dots,\ell-1.
\]
Here $x_{k}$ is a $b\times n_{k}$ matrix of input/hidden/output
units, $d_{k}$ is a $b\times n_{k}$ dropout-mask matrix of independent
Bernoulli($1-p_{k}$) random variables, $p_{k}$ denotes the probability
of dropping out units in level $k$, and $W_{k}$ is an $n_{k}\times n_{k+1}$
matrix of weights connecting level $k$ with level $k+1$. We are
using $\cdot$ for (Hadamard) element-wise multiplication and $\times$
for matrix multiplication. We have forgotten to include non-linear
functions (e.g. the rectifier function for the hidden units, and softmax
for the output units) but for the introduction we will keep the network
as simple as possible.

The network can be trained using the backpropagation algorithm to
calculate the gradients of a cost function (e.g. negative log-likelihood)
with respect to the $W_{k}$:
\begin{align*}
\frac{\partial\mathrm{cost}}{\partial W_{k}} & =[x_{k}\cdot d_{k}]^{\mathsf{T}}\times\frac{\partial\mathrm{cost}}{\partial x_{k+1}}\\
\frac{\partial\mathrm{cost}}{\partial x_{k}} & =\left(\frac{\partial\mathrm{cost}}{\partial x_{k+1}}\times W_{k}^{\mathrm{\mathsf{T}}}\right)\cdot d_{k}.
\end{align*}
With dropout training, we are trying to minimize the cost function
averaged over an ensemble of closely related networks. However, networks
typically contain thousands of hidden units, so the size of the ensemble
is \emph{much} larger than the number of training samples that can
possibly be `seen' during training. This suggests that the independence
of the rows of the dropout mask matrices $d_{k}$ might not be terribly
important; the success of dropout simply cannot depend on exploring
a large fraction of the available dropout masks. Some machine learning
libraries such as Pylearn2 allow dropout to be applied batchwise instead
of independently%
\footnote{\href{https://github.com/lisa-lab/pylearn2/blob/master/pylearn2/models/mlp.py}{Pylearn2: see function apply\_{}dropout in mlp.py}%
}. This is done by replacing $d_{k}$ with a $1\times n_{k}$ row matrix
of independent Bernoulli$(1-p_{k})$ random variables, and then copying
it vertically $b$ times to get the right shape.

To be practical, it is important that each training minibatch can
be processed quickly. A crude way of estimating the processing time
is to count the number of floating point multiplication operations
needed (naively) to evaluate the $\times$ matrix multiplications
specified above:
\[
\sum_{k=0}^{\ell-1}\underbrace{b\times n_{k}\times n_{k+1}}_{\mathrm{forwards}}+\underbrace{n_{k}\times b\times n_{k+1}}_{\partial\mathrm{cost/\partial}W}+\underbrace{b\times n_{k+1}\times n_{k}}_{\mathrm{backwards}}.
\]
However, when we take into account the effect of the dropout mask,
we see that many of these multiplications are unnecessary. The $(i,j)$-th
element of the $W_{k}$ weight matrix effectively `drops-out' of
the calculations if unit $i$ is dropped in level $k$, or if unit
$j$ is dropped in level $k+1$. Applying 50\% dropout in levels $k$
and $k+1$ renders 75\% of the multiplications unnecessary.

If we apply dropout independently, then the parts of $W_{k}$ that
disappear are different for each sample. This makes it effectively
impossible to take advantage of the redundancy---it is slower to check
if a multiplication is necessary than to just do the multiplication.
However, if we apply dropout batchwise, then it becomes easy to take
advantage of the redundancy. We can literally drop-out redundant parts
of the calculations.

\begin{figure*}
\begin{centering}
\includegraphics[width=0.38\textwidth]{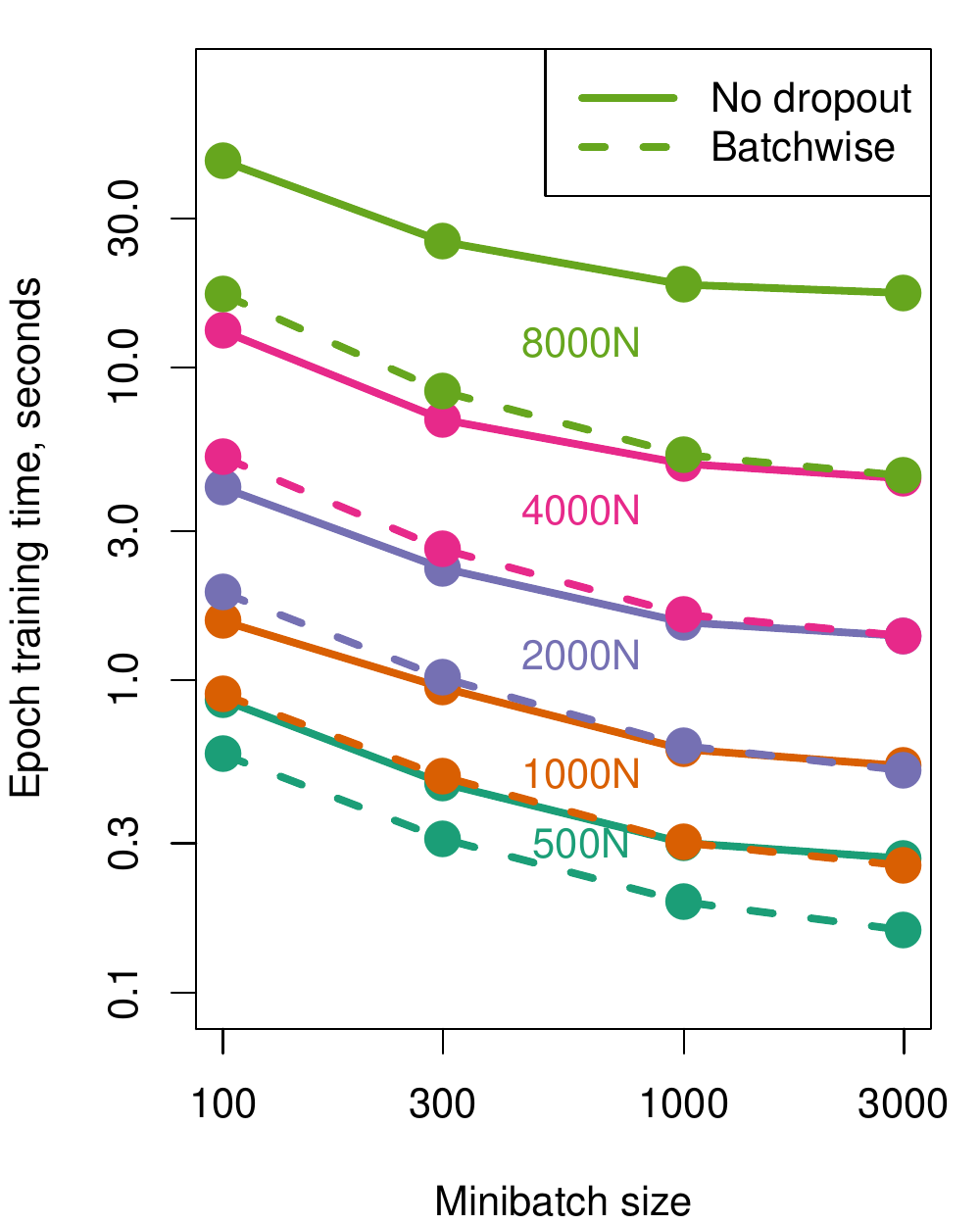}\includegraphics[width=0.38\textwidth]{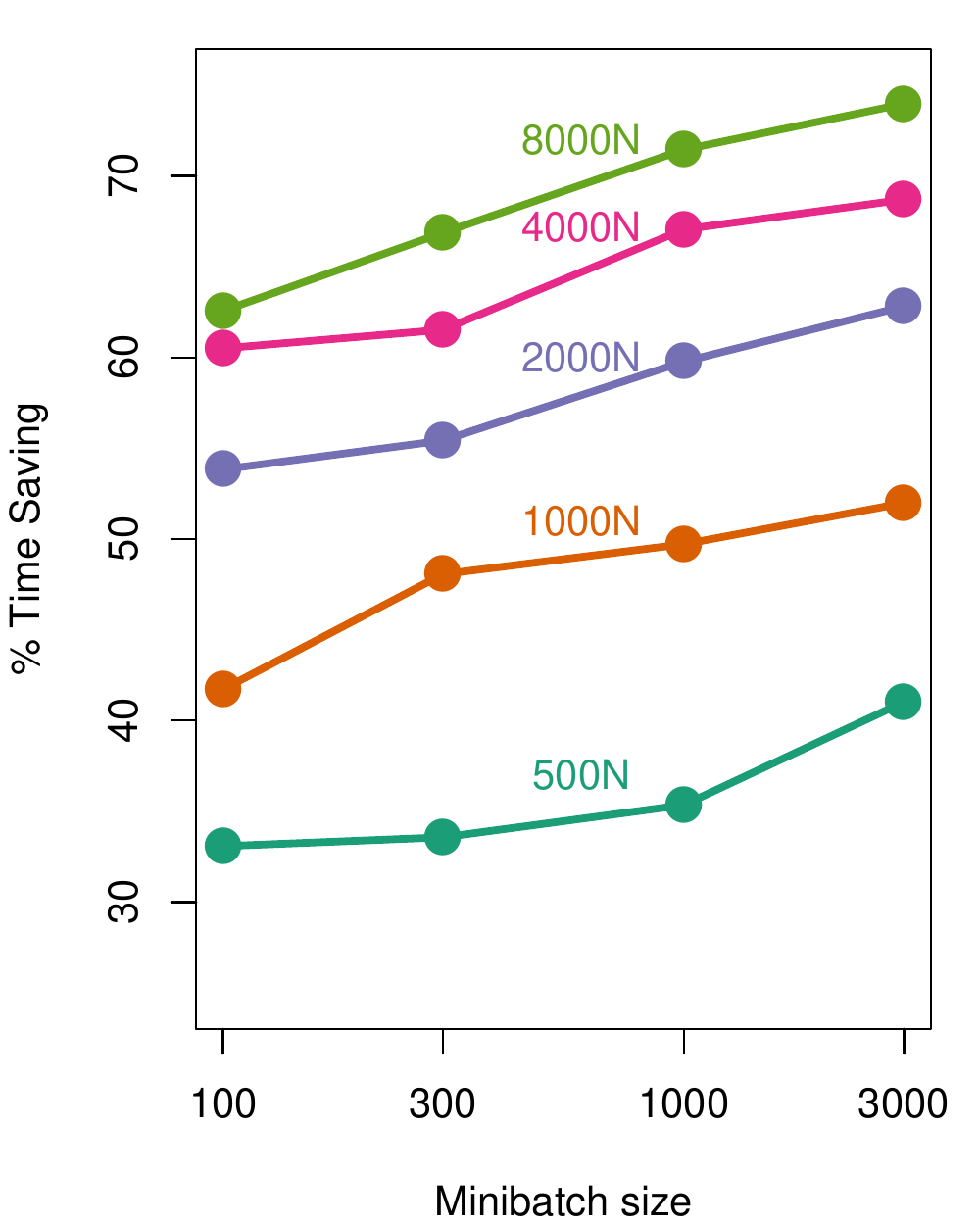}
\par\end{centering}
\caption{Left: MNIST training time for three layer networks (log scales) on
an NVIDIA GeForce GTX 780 graphics card. Right: Percentage reduction
in training times moving from no dropout to batchwise dropout. The time saving for the 500N network with minibatches of size 100 increases from 33\% to 42\% if you instead compare batchwise dropout with independent dropout.\label{fig:Training-time}}
\end{figure*}

The binary $1\times n_{k}$ batchwise dropout matrices $d_{k}$
naturally define submatrices of the weight and hidden-unit matrices.
Let $x_{k}^{\mathrm{dropout}}:=x_{k}[\,:\,,d_{k}]$ denote the submatrix
of $x_{k}$ consisting of the level-$k$ hidden units that survive
dropout. Let $W_{k}^{\mathrm{dropout}}:=W_{k}[d_{k},d_{k+1}]$ denote
the submatrix of $W_{k}$ consisting of weights that connect active
units in level $k$ to active units in level $k+1$. The network can
then be trained using the equations:
\begin{align*}
x_{k+1}^{\mathrm{dropout}} & =x_{k}^{\mathrm{dropout}}\times W_{k}^{\mathrm{dropout}}\\
\frac{\partial\mathrm{cost}}{\partial W_{k}^{\mathrm{dropout}}} & =(x_{k}^{\mathrm{dropout}})^{\mathsf{T}}\times\frac{\partial\mathrm{cost}}{\partial x_{k+1}^{\mathrm{dropout}}}\\
\frac{\partial\mathrm{cost}}{\partial x_{k}^{\mathrm{dropout}}} & =\frac{\partial\mathrm{cost}}{\partial x_{k+1}^{\mathrm{dropout}}}\times(W_{k}^{\mathrm{dropout}})^{\mathsf{T}}
\end{align*}
The redundant multiplications have been eliminated. There is an additional
benefit in terms of memory needed to store the hidden units: $x_{k}^{\mathrm{dropout}}$
needs less space than $x_{k}$. In Section \ref{sec:Implementation}
we look at the performance improvement that can be achieved using
CUDA/CUBLAS code running on a GPU. Roughly speaking, processing a
minibatch with 50\% batchwise dropout takes as long as training a
50\% smaller network on the same data. This explains the nearly overlapping
pairs of lines in Figure \ref{fig:Training-time}.

We should emphasize that batchwise dropout only improves performance
during training; during testing the full $W_{k}$ matrix is used as
normal, scaled by a factor of $1-p_k$. However, machine learning
research is often constrained by long training times and high costs
of equipment. In Section \ref{sec:Results-for-fully-connected} we
show that all other things being equal, batchwise dropout is similar
to independent dropout, but faster. Moreover, with the increase in
speed, all other things do not have to be equal. With the same resources,
batchwise dropout can be used to
\begin{itemize}
\item increase the number of training epochs,
\item increase the number of hidden units,
\item increase the number of validation runs used to optimize ``hyper-parameters'',
or
\item to train a number of independent copies of the network to form a committee.
\end{itemize}
These possibilities will often be useful as ways of improving generalization/reducing
test error.

In Section \ref{sec:Convolutional-networks} we look at batchwise
dropout for convolutional networks. Dropout for convolutional networks
is more complicated as weights are shared across spatial locations.
A minibatch passing up through a convolutional network might be represented
at an intermediate hidden layer by an array of size $100\times32\times12\times12$:
100 samples, the output of 32 convolutional filters, at each of $12\times12$
spatial locations. It is conventional to use a dropout mask with shape
$100\times32\times12\times12$; we will call this independent dropout.
In contrast, if we want to apply batchwise dropout efficiently by
adapting the submatrix trick, then we will effectively be using a
dropout mask with shape $1\times32\times1\times1$. This looks like
a significant change: we are modifying the ensemble over which the
average cost is optimized. During training, the error rates are higher.
However, testing the networks gives very similar error rates.

\subsection{Fast dropout}

We might have called batchwise dropout \emph{fast dropout }but that
name is already taken\emph{ }\cite{FastDropout}. Fast dropout is
very different approach to solving the problem of training large neural
network quickly without overfitting. We discuss some of the differences
of the two techniques in the appendix.

\section{Implementation\label{sec:Implementation}}

In theory, for $n\times n$ matrices, addition is an $\mathrm{O}(n^{2})$
operation, and multiplication is $\mathrm{O}(n^{2.37...})$ by the
Coppersmith\textendash{}Winograd algorithm. This suggests that the
bulk of our processing time should be spent doing matrix multiplication,
and that a performance improvement of about 60\% should be possible
compared to networks using independent dropout, or no dropout at all.
In practice, SGEMM functions use Strassen's algorithm or naive matrix
multiplication, so performance improvement of up to 75\% should be
possible.

We implemented batchwise dropout for fully-connected and convolutional
neural networks using CUDA/CUBLAS\footnote{Software available at \href{http://www2.warwick.ac.uk/fac/sci/statistics/staff/academic-research/graham/dropout.zip}{http://www2.warwick.ac.uk/fac/sci/statistics/staff/academic-research/graham/}}. We found that using the highly optimized \textsf{cublasSgemm} function
to do the bulk of the work, with CUDA kernels used to form the submatrices
$W_{k}^{\mathrm{dropout}}$ and to update the $W_{k}$ using $\partial\mathrm{cost}/\partial W_{k}^{\mathrm{dropout}}$,
worked well. Better performance may well be obtained by writing a
SGEMM-like matrix multiplication function that understands submatrices.

For large networks and minibatches, we found that batchwise dropout
was substantially faster, see Figure \ref{fig:Training-time}. The
approximate overlap of some of the lines on the left indicates that 50\% batchwise
dropout reduces the training time in a similar manner to halving the
number of hidden units.

The graph on the right show the time saving obtained by using submatrices to implement dropout.
Note that for consistency with the left hand side, the graph compares batchwise dropout with dropout-free
networks, \emph{not} with networks using independent dropout. The
need  to implement dropout masks for independent dropout means that
Figure 1 slightly undersells the performance benefits of batchwise
dropout as an alternative to independent dropout.

For smaller networks, the performance improvement is lower---bandwidth
issues result in the GPU being under utilized. If you were implementing
batchwise dropout for CPUs, you would expect to see greater performance
gains for smaller networks as CPUs have a lower processing-power to
bandwidth ratio.

\subsection{Efficiency tweaks \label{sub:Fixed-dropout-amounts}}

If you have $n=2000$ hidden units and you drop out $p=50$\% of them,
then the number of dropped units is approximately $np=1000$, but
with some small variation as you are really dealing with a Binomial$(n,p)$
random variable---its standard deviation is $\sqrt{np(1-p)}=22.4$.
The sizes of the submatrices $W_{k}^{\mathrm{dropout}}$ and $x_{k}^{\mathrm{dropout}}$
are therefore slightly random. In the interests of efficiency and
simplicity, it is convenient to remove this randomness. An alternative
to dropping each unit independently with probability $p$ is to drop
a subset of exactly $np$ of the hidden units, uniformly at random
from the set of all $\binom{n}{np}$ such subsets. It is still the
case that each unit is dropped out with probability $p$. However,
within a hidden layer we no longer have strict independence regarding
which units are dropped out. The probability of dropping out the first
two hidden units changes very slightly, from
\[
p^{2}=0.25\qquad\mathrm{to}\qquad\frac{np}{n}\cdot\frac{np-1}{n-1}=0.24987....
\]
Also, we used a modified form of NAG-momentum minibatch gradient descent \cite{conf/icml/SutskeverMDH13}.
After each minibatch, we only updated the elements of $W_{k}^{\mathrm{dropout}}$,
not all the element of $W_{k}$. With $v_{k}$ and $v_{k}^{\mathrm{dropout}}$
denoting the momentum matrix/submatrix corresponding to $W_{k}$ and
$W_{k}^{\mathrm{dropout}}$, our update was
\begin{align*}
v_{k}^{\mathrm{dropout}} & \leftarrow\mu v_{k}^{\mathrm{dropout}}-\varepsilon(1-\mu)\partial\mathrm{cost}/\partial W_{k}^{\mathrm{dropout}}\\
W_{k}^{\mathrm{dropout}} & \leftarrow W_{k}^{\mathrm{dropout}}+v_{k}^{\mathrm{dropout}}.
\end{align*}
The momentum still functions as an autoregressive process, smoothing
out the gradients, we are just reducing the rate of decay $\mu$ by
a factor of $(1-p_{k})(1-p_{k+1})$.

\begin{figure}[t]
\begin{centering}
\includegraphics[width=0.45\textwidth]{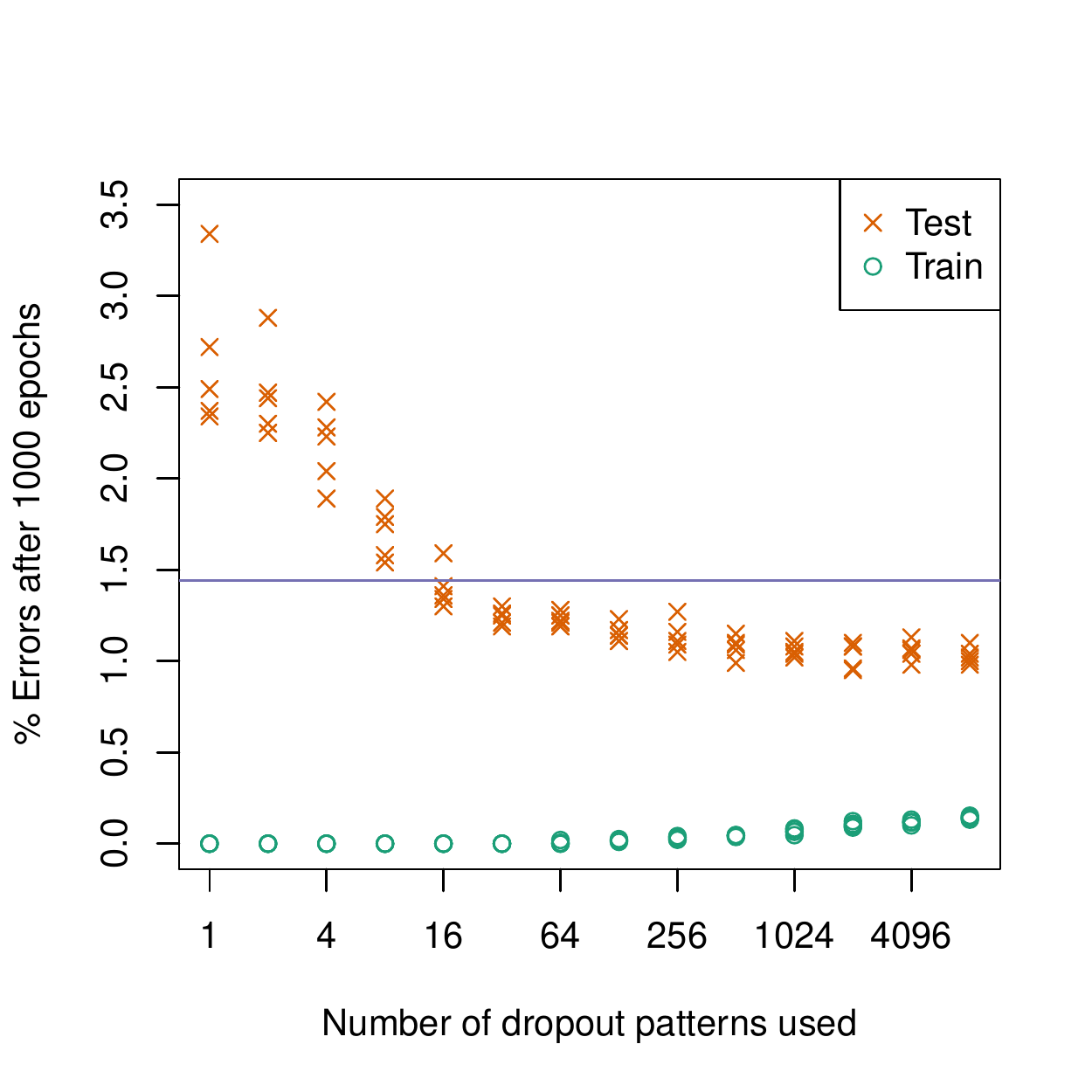}
\par\end{centering}
\caption{\label{fig:limited-dropout-patterns}
Dropout networks trained using a restricted the number of dropout patterns (each $\times$ is from an independent experiment).
The blue line marks the test error for a network with half as many hidden units trained without dropout.
}
\end{figure}

\section{Results for fully-connected networks\label{sec:Results-for-fully-connected}}

The fact that batchwise dropout takes less time per training epoch
would count for nothing if a much larger number of epochs was needed
to train the network, or if a large number of validation runs were
needed to optimize the training process. We have carried out a number
of simple experiment to compare independent and batchwise dropout.
In many cases we could have produced better results by increasing
the training time, annealing the learning rate, using validation to
adjust the learning process, etc. We choose not to do this as the
primary motivation for batchwise dropout is efficiency, and excessive
use of fine-tuning is not efficient.

For datasets, we used:
\begin{itemize}
\item The MNIST%
\footnote{http://yann.lecun.com/exdb/mnist/%
} set of $28\times28$ pixel handwritten digits.
\item The CIFAR-10 dataset of 32x32 pixel color pictures (\cite{CIFAR10}).
\item An artificial dataset designed to be easy to overfit.
\end{itemize}
Following \cite{dropout}, for MNIST and CIFAR-10 we trained networks
with 20\% dropout in the input layer, and 50\% dropout in the hidden
layers. For the artificial dataset we increased the input-layer dropout
to 50\% as this reduced the test error. In some cases, we have used
relatively small networks so that we would have time to train a number
of independent copies of the networks. This was useful in order to
see if the apparent differences between batchwise and independent
dropout are significant or just noise.

\subsection{MNIST \label{sub:MNIST-using-only}}

Our first experiment explores the effect of dramatically restricting
the number of dropout patterns seen during training. Consider a network
with three hidden layers of size 1000, trained for 1000 epochs using
minibatches of size 100. The number of distinct dropout patterns,
$2^{3784}$, is so large that we can assume that we will never generate
the same dropout mask twice. During independent dropout training we
will see 60 million different dropout patterns, during batchwise dropout
training we will see 100 times fewer dropout patterns.

For both types of dropout, we trained 12 independent networks for
1000 epochs, with batches of size 100. For batchwise dropout we got
a mean test error of 1.04\% {[}range (0.92\%,1.1\%), s.d. 0.057\%{]}
and for independent dropout we got a mean test errors of 1.03\% {[}range
(0.98\%,1.08\%), s.d. 0.033\%{]}. The difference in the mean test
errors is not statistically significant.

To explore further the reduction in the number of dropout patterns
seen, we changed our code for (pseudo)randomly generating batchwise
dropout patterns to restrict the number of distinct dropout patterns
used. We modified it to have period $n$ minibatches, with $n=1,2,4,8,\dots$;
see Figure \ref{fig:limited-dropout-patterns}. For $n=1$ this corresponds
to only ever using one dropout mask, so that 50\% of the network's
3000 hidden weights are never actually trained (and 20\% of the 784
input features are ignored). During training this corresponds to training
a dropout-free network with half as many hidden units---the test error for such a network is marked by a blue line in Figure \ref{fig:limited-dropout-patterns}.
The error during testing is higher
than the blue line because the untrained weights add noise to the network.

If $n$ is less than thirteen, is it likely that some of the networks
3000 hidden units are dropped out every time and so receive no training.
If $n$ is in the range thirteen to fifty, then it is likely that
every hidden unit receives some training, but some pairs of hidden
units in adjacent layers will not get the chance to interact during
training, so the corresponding connection weight is untrained. As
the number of dropout masks increases into the hundreds, we see that
it is quickly a case of diminishing returns.

\begin{figure*}[t]
\begin{centering}
\includegraphics[width=0.4\textwidth]{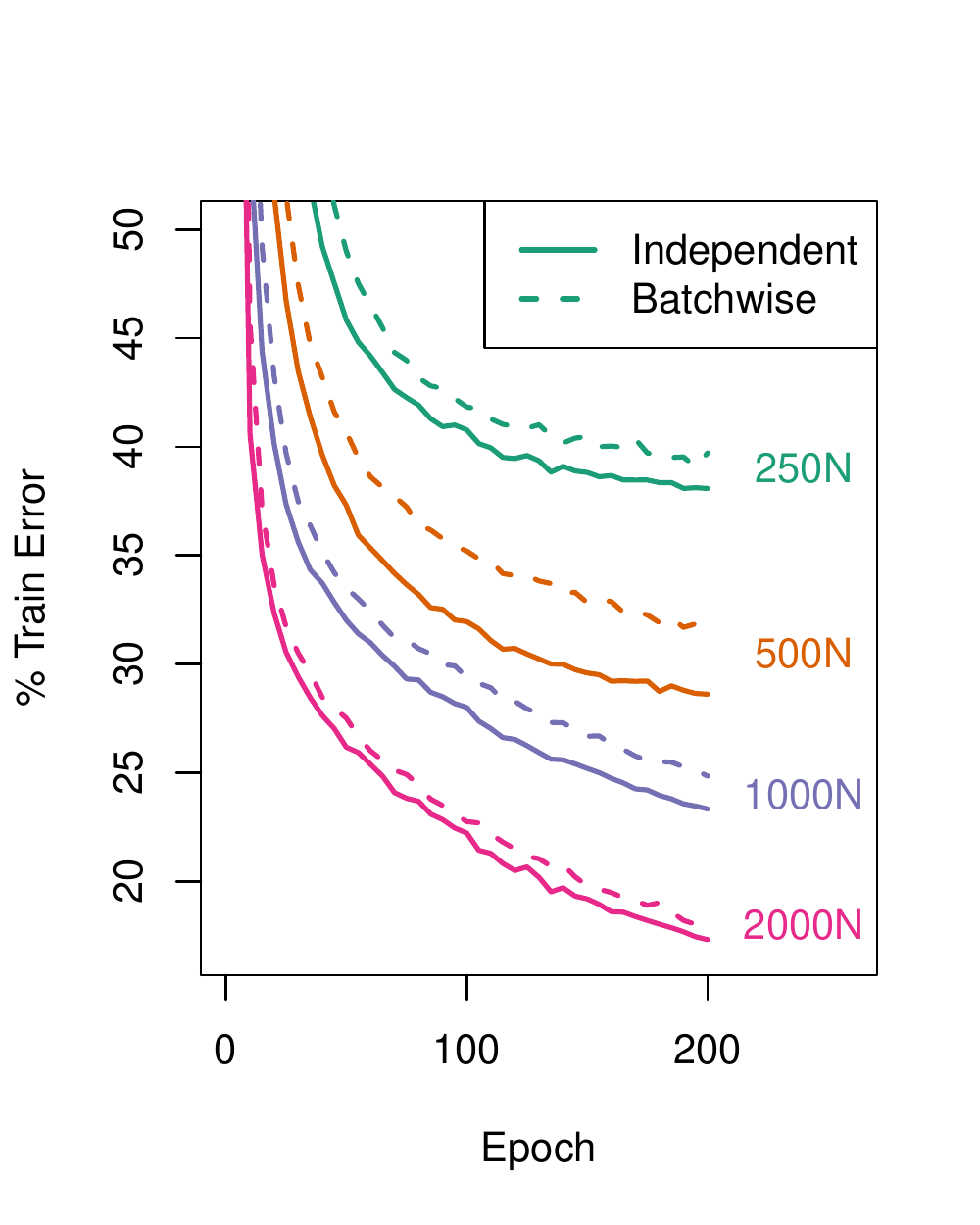}\includegraphics[width=0.4\textwidth]{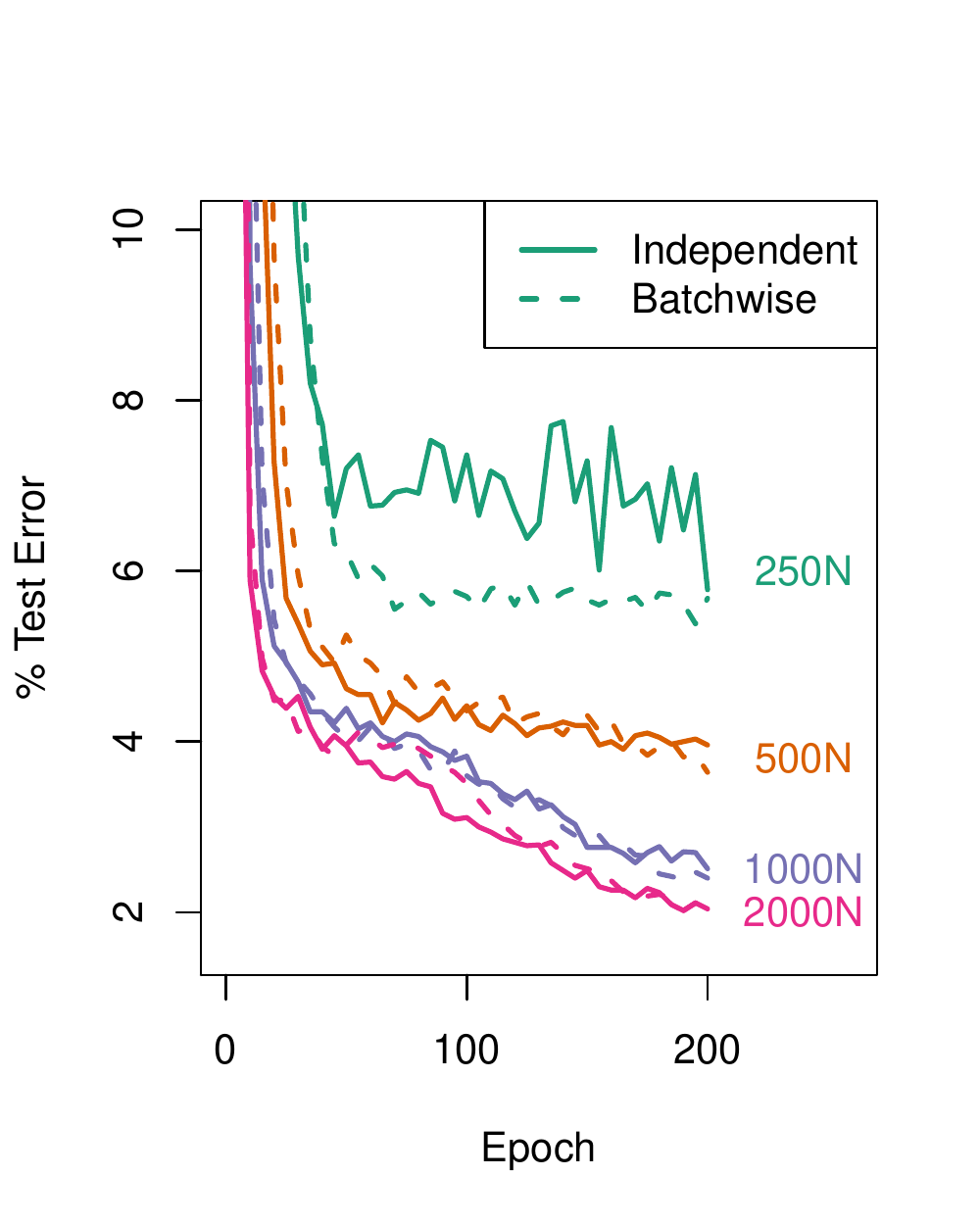}
\par\end{centering}

\caption{Artificial dataset. 100 classes each corresponding to noisy observations
of a one dimensional manifold in $\{0,1\}^{1000}$. \label{fig:Artificial-dataset.-100}}
\end{figure*}

\subsection{Artificial dataset}

To test the effect of changing network size, we created an artificial
dataset. It has 100 classes, each containing 1000 training samples and 100 test samples.
Each class is defined using an independent random walk of length 1000 in the discrete cube $\{0,1\}^{1000}$. For each class we generated the random walk, and then used it to produce the training and test samples by randomly picking points along the length of walk (giving binary
sequences of length 1000) and then randomly flipping 40\% of the bits.
We trained three layer networks with $n\in\{250,500,1000,2000\}$
hidden units per layer with minibatches of size 100. See Figure~\ref{fig:Artificial-dataset.-100}.

Looking at the training error against training epochs, independent dropout seems
to learn slightly faster. However, looking at the test errors over time, there does not seem to be much difference between the two forms of dropout. Note that the
$x$-axis is the number of training epochs, not the training time. The batchwise dropout networks are learning much faster in terms of real time.

\begin{figure*}[t]
\begin{centering}
\includegraphics[width=0.4\textwidth]{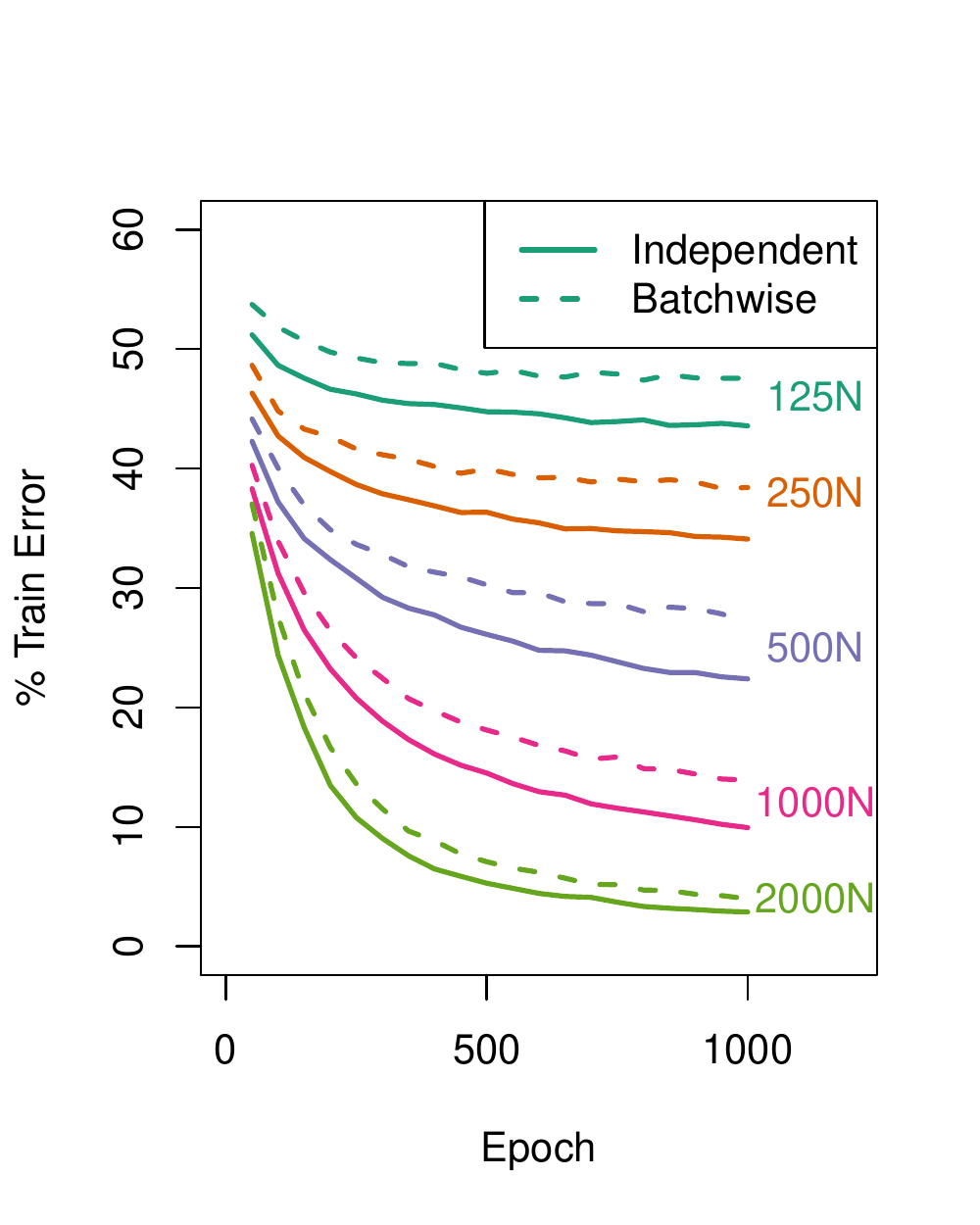}\includegraphics[width=0.4\textwidth]{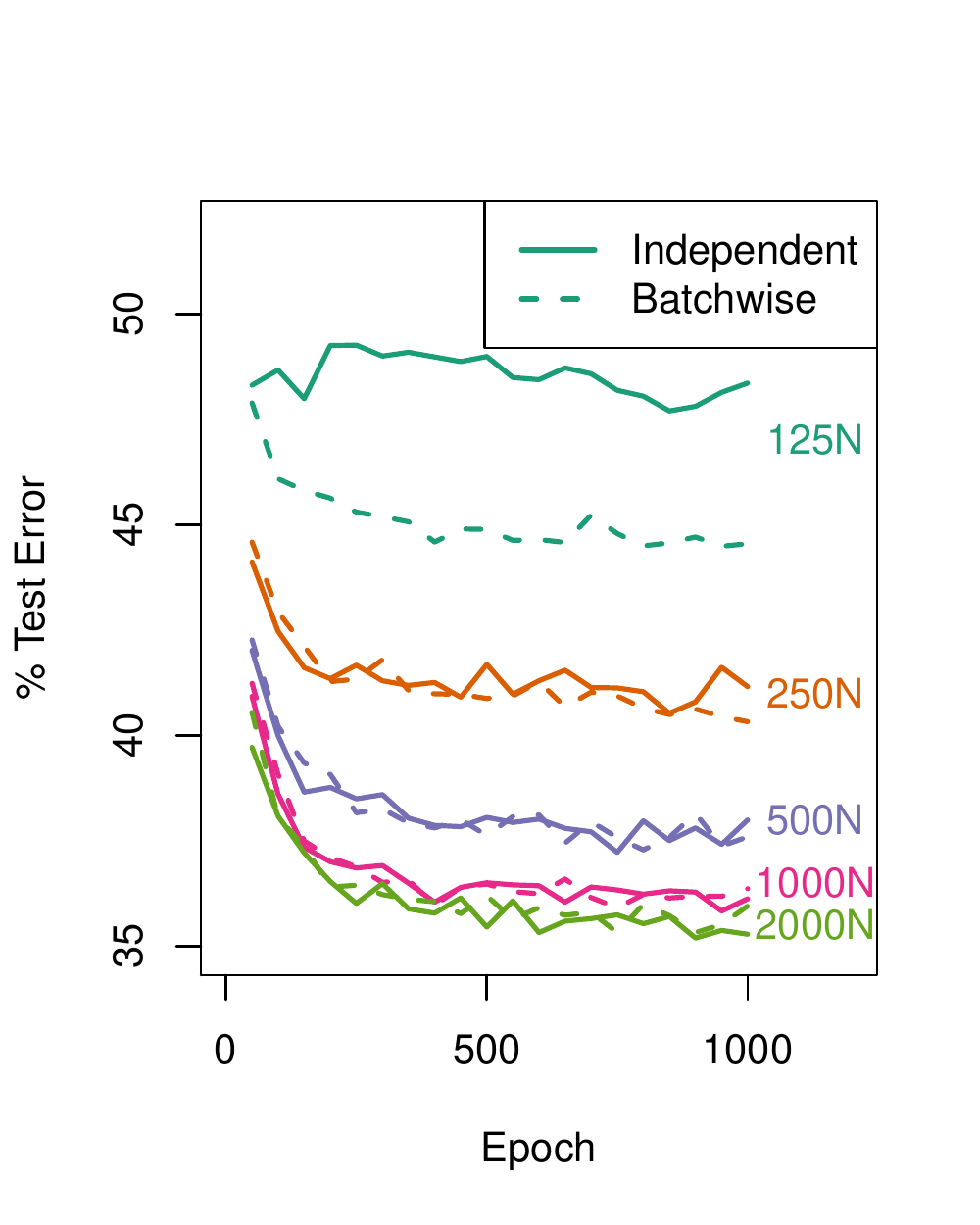}
\par\end{centering}

\caption{Results for CIFAR-10 using fully-connected networks of different sizes.\label{fig:cifar-fc}}
\end{figure*}

\subsection{CIFAR-10 fully-connected\label{sub:CIFAR-10-fully-connected}}

Learning CIFAR-10 using a fully connected network is rather difficult.
We trained three layer networks with $n\in\{125,250,500,1000,2000\}$
hidden units per layer with minibatches of size 1000. We augmented the training data with horizontal flips. See Figure \ref{fig:cifar-fc}.

\section{Convolutional networks\label{sec:Convolutional-networks}}

Dropout for convolutional networks is more complicated as weights
are shared across spatial locations. Suppose layer $k$ has spatial
size $s_{k}\times s_{k}$ with $n_{k}$ features per spatial location,
and if the $k$-th operation is a convolution with $f\times f$ filters.
For a minibatch of size $b$, the convolution involves arrays with
sizes:
\begin{align*}
\mathrm{layer}\ k: & \ b\times n_{k}\times s_{k}\times s_{k}\\
\mathrm{weights\ }W_{k}: & \ n_{k+1}\times n_{k}\times f\times f
\end{align*}
Dropout is normally applied using dropout masks with the same size
as the layers. We will call this independent dropout---independent
decisions are mode at every spatial location. In contrast, we define
batchwise dropout to mean using a dropout mask with shape $1\times n_{k}\times1\times1$.
Each minibatch, each convolutional filter is either on or off---across
all spatial locations.

These two forms of regularization seem to be doing quite different
things. Consider a filter that detects the color red, and a picture
with a red truck in it. If dropout is applied independently, then
by the law of averages the message ``red'' will be transmitted with
very high probability, but with some loss of spatial information.
In contrast, with batchwise dropout there is a 50\% chance we delete
the entire filter output. Experimentally, the only substantial difference
we could detect was that batchwise dropout resulted in larger errors during training.

To implement batchwise dropout efficiently, notice that the $1\times n_{k}\times1\times1$
dropout masks corresponds to forming subarrays $W_{k}^{\mathrm{dropout}}$
of the weight arrays $W_{k}$ with size
\[
(1-p_{k+1})n_{k+1}\times(1-p_{k})n_{k}\times f\times f.
\]
The forward-pass is then simply a regular convolutional operation
using $W_{k}^{\mathrm{dropout}}$; that makes it possible, for example,
to take advantage of the highly optimized $\mathsf{cudnnConvolutionForward}$
function from the \href{https://developer.nvidia.com/cuDNN}{NVIDIA cuDNN}
package.

\subsection{MNIST}

For MNIST, we trained a LeNet-5 type CNN with two layers of $5\times5$
filters, two layers of $2\times2$ max-pooling, and a fully connected
layer \cite{Lenet}. There are three places for applying 50\% dropout:
\[
32\mathrm{C}5-\mathrm{MP2}\overset{50\%}{-}64\mathrm{C}5-\mathrm{MP2}\overset{50\%}{-}512\mathrm{N}\overset{50\%}{-}\mathrm{10N}.
\]
The test errors for the two dropout methods are similar, see Figure
\ref{fig:MNIST-test-errors,}.

\begin{figure}[t]
\begin{centering}
\includegraphics[width=0.38\textwidth]{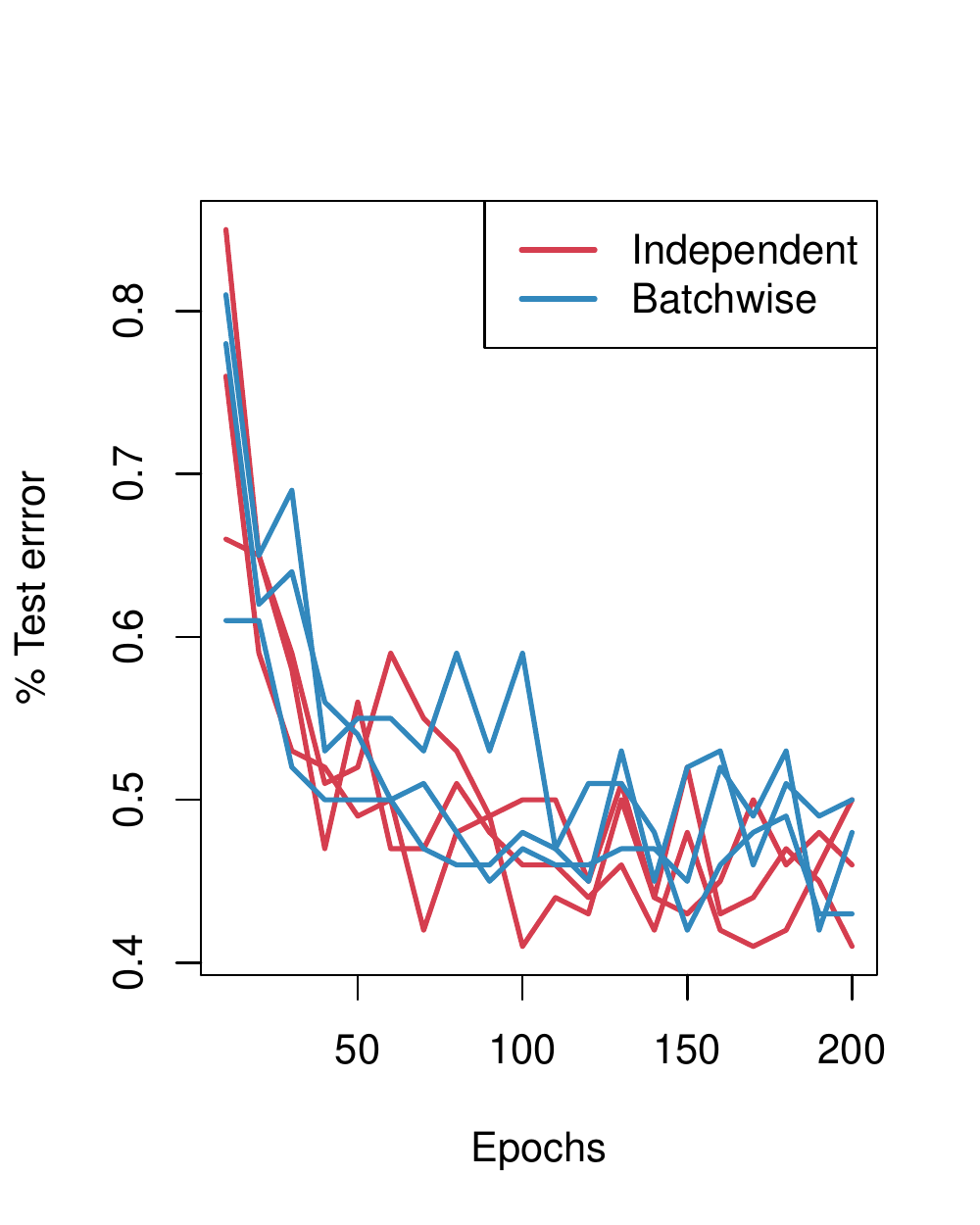}
\par\end{centering}

\centering{}\caption{MNIST test errors, training repeated three times for both dropout
methods.\label{fig:MNIST-test-errors,}}
\end{figure}

\subsection{CIFAR-10 with varying dropout intensity\label{sub:Shallow-CIFAR-10}}

For a first experiment with CIFAR-10 we used a small convolutional network with small filters.
The network is a scaled down version of the network from \cite{multicolumndeep}; there are four places to apply dropout:
\[
128\mathrm{C}3-\mathrm{MP2}\overset{p}{-}256\mathrm{C2}-\mathrm{MP2}\overset{p}{-}384\mathrm{C2}-\mathrm{MP2}\overset{p}{-}512\mathrm{N}\overset{p}{-}\mathrm{10N}.
\]
The input layer is $24\times 24$. We trained the network for 1000 epochs using randomly chosen subsets of the training images, and reflected each image horizontally with probability one half. For testing we used the centers of the images.

\begin{figure}[t]
\begin{centering}
\includegraphics[width=0.45\textwidth]{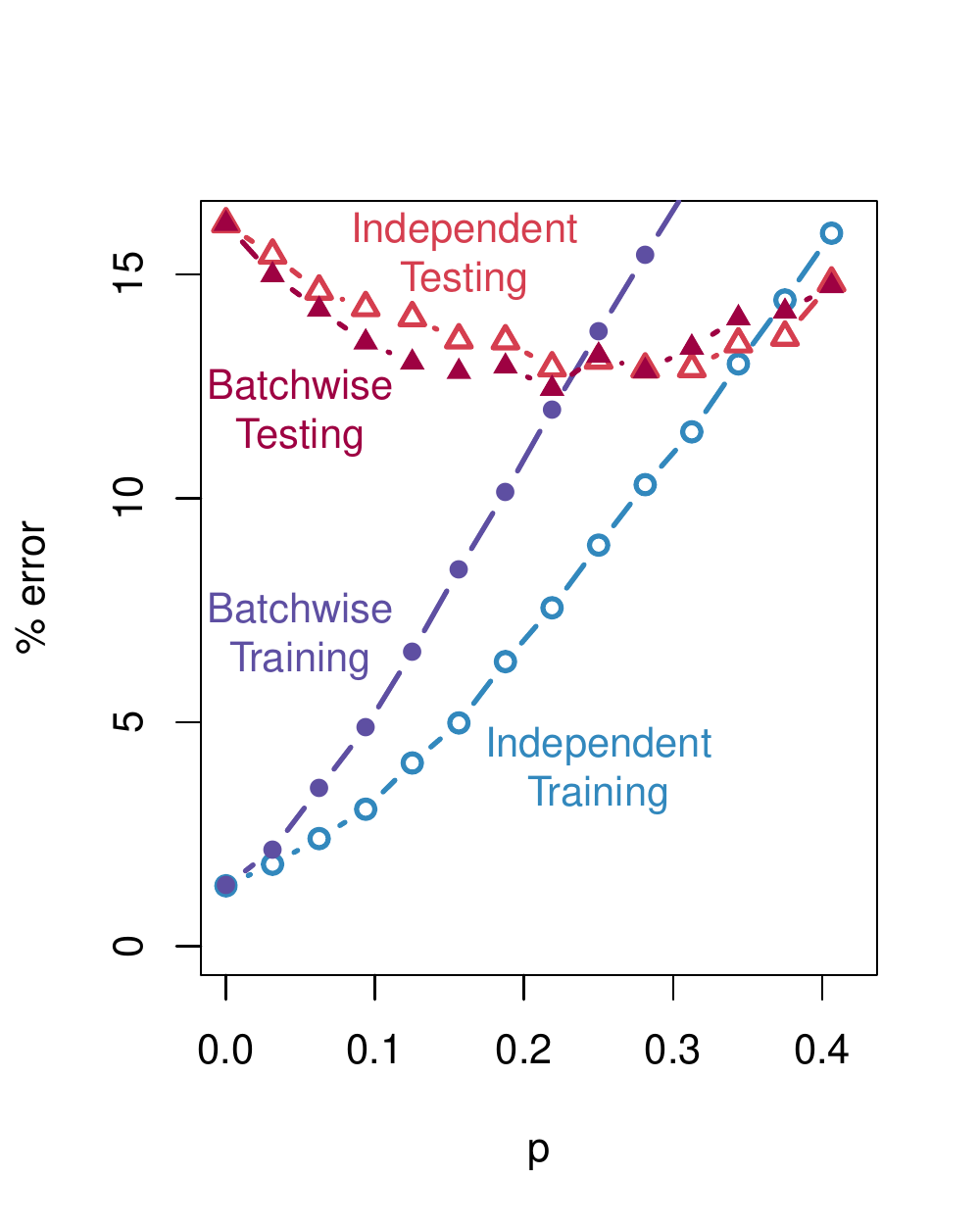}
\par\end{centering}

\caption{CIFAR-10 results using a convolutional network with dropout probability
$p\in(0,0.4)$. Batchwise dropout produces a slightly
lower minimum test error. \label{fig:CIFAR-10-results-using}}
\end{figure}

In Figure \ref{fig:CIFAR-10-results-using} we show the effect of
varying the dropout probability $p$. The training errors are
increasing with $p$, and the training errors are higher for batchwise dropout.
The test-error curves both seem to have local minima around $p=0.2$. The batchwise test error curve seems to be shifted slightly to the left of the independent one, suggesting that for any given value of $p$, batchwise dropout is a slightly stronger form of regularization.

\subsection{CIFAR-10 with many convolutional layers\label{sub:Deeper-CIFAR-10}}

We trained a deep convolutional network on CIFAR-10 \emph{without}
data augmentation. Using the notation of \cite{FractionalMaxPooling},
our network has the form
\[
(64nC2-FMP\sqrt[3]{2})_{12}-832C2-896C1-\mathrm{output},
\]
i.e. it consists of 12 $2\times2$ convolutions with $64n$ filters in the $n$-th layer,
12 layers max-pooling, followed by two fully connected layers; the network has 12.6 million
parameters. We used an increasing amount of dropout per layer, rising
linearly from 0\% dropout after the third layer to 50\% dropout after
the 14th. Even though the amount of dropout used in the middle layers
is small, batchwise dropout took less than half as long per epoch
as independent dropout; this is because applying small amounts of
independent dropout in large hidden-layers creates a bandwidth performance-bottleneck.

As the network's max-pooling operation is stochastic, the test errors can be reduced by repetition. Batchwise dropout resulted in a average test error of 7.70\% (down to 5.78\% with 12-fold testing). Independent
dropout resulted in an average test error of 7.63\% (reduced to 5.67\% with 12-fold testing).

\section{Conclusions and future work}

We have implemented an efficient form of batchwise dropout. All other
things being equal, it seems to learn at roughly the same speed as independent dropout, but each epoch is faster. Given a fixed computational budget, it will often
allow you to train better networks.

There are other potential uses for batchwise dropout that we have
not explored yet:
\begin{itemize}
\item Restricted Boltzmann Machines can be trained by contrastive divergence \cite{HinSal06} with dropout \cite{dropout}. Batchwise dropout could be used to increase the speed of training.
\item When a fully connected network sits on top of a convolutional network,
training the top and bottom of the network can be separated over different
computational nodes \cite{OneWeirdTrick}. The fully connected top-parts
of the network typically contains 95\% of the parameters---keeping
the nodes synchronized is difficult due to the large size of the matrices.
With batchwise dropout, nodes could communicate $\partial\mathrm{cost}/\partial W_{k}^{\mathrm{dropout}}$
instead of $\partial\mathrm{cost}/\partial W_{k}$ and so reducing
the bandwidth needed.
\item Using independent dropout with recurrent neural networks can be too
disruptive to allow effective learning; one solution is to only apply
dropout to some parts of the network \cite{RecuurentNetworkRegularization}.
Batchwise dropout may provide a less damaging form of dropout, as
each unit will either be on or off for the whole time period.
\item Dropout is normally only used during training. It is generally more
accurate use the whole network for testing purposes; this is equivalent
to averaging over the ensemble of dropout patterns. However, in a
``real-time'' setting, such as analyzing successive frames from
a video camera, it may be more efficient to use dropout during testing,
and then to average the output of the network over time.
\item Nested dropout \cite{NestedDropout} is a variant of regular dropout
that extends some of the properties of PCA to deep networks. Batchwise
nested dropout is particularly easy to implement as the submatrices
are regular enough to qualify as matrices in the context of the SGEMM
function (using the LDA argument).
\item DropConnect is an alternative form of regularization to dropout \cite{DropConnect}.
Instead of dropping hidden units, individual elements of the weight
matrix are dropped out. Using a modification similar to the one in
Section \ref{sub:Fixed-dropout-amounts}, there are opportunities
for speeding up DropConnect training by approximately a factor of
two.
\end{itemize}


\begin{thebibliography}{10}

\bibitem{multicolumndeep}
D.~Ciresan, U.~Meier, and J.~Schmidhuber.
\newblock \href{www.idsia.ch/~juergen/cvpr2012.pd}{Multi-column deep neural
  networks for image classification}.
\newblock In {\em Computer Vision and Pattern Recognition (CVPR), 2012 IEEE
  Conference on}, pages 3642--3649, 2012.

\bibitem{FractionalMaxPooling}
Ben Graham.
\newblock Fractional max-pooling, 2014.
\newblock
  \href{http://arxiv.org/abs/1412.6071}{http://arxiv.org/abs/1412.6071}.

\bibitem{HinSal06}
Hinton and Salakhutdinov.
\newblock \href{http://www.cs.toronto.edu/~hinton/science.pdf}{Reducing the
  Dimensionality of Data with Neural Networks}.
\newblock {\em SCIENCE: Science}, 313, 2006.

\bibitem{CIFAR10}
Alex Krizhevsky.
\newblock \href{http://www.cs.toronto.edu/~kriz/cifar.html}{Learning Multiple
  Layers of Features from Tiny Images}.
\newblock Technical report, 2009.

\bibitem{OneWeirdTrick}
Alex Krizhevsky.
\newblock One weird trick for parallelizing convolutional neural networks,
  2014.
\newblock
  \href{http://arxiv.org/abs/1404.5997}{http://arxiv.org/abs/1404.5997}.

\bibitem{Lenet}
Y.~L. Le~Cun, L.~Bottou, Y.~Bengio, and P.~Haffner.
\newblock Gradient-based learning applied to document recognition.
\newblock {\em Proceedings of IEEE}, 86(11):2278--2324, November 1998.

\bibitem{NestedDropout}
Oren Rippel, Michael~A. Gelbart, and Ryan~P. Adams.
\newblock Learning ordered representations with nested dropout, 2014.
\newblock
  \href{http://arxiv.org/abs/1402.0915}{http://arxiv.org/abs/1402.0915}.

\bibitem{dropout}
Nitish Srivastava, Geoffrey Hinton, Alex Krizhevsky, Ilya Sutskever, and Ruslan
  Salakhutdinov.
\newblock \href{http://jmlr.org/papers/v15/srivastava14a.html}{Dropout: A
  Simple Way to Prevent Neural Networks from Overfitting}.
\newblock {\em Journal of Machine Learning Research}, 15:1929--1958, 2014.

\bibitem{conf/icml/SutskeverMDH13}
Ilya Sutskever, James Martens, George~E. Dahl, and Geoffrey~E. Hinton.
\newblock \href{http://jmlr.org/proceedings/papers/v28/}{On the importance of
  initialization and momentum in deep learning}.
\newblock In {\em ICML}, volume~28 of {\em JMLR Proceedings}, pages 1139--1147.
  JMLR.org, 2013.

\bibitem{DropConnect}
Li~Wan, Matthew Zeiler, Sixin Zhang, Yann Lecun, and Rob Fergus.
\newblock
  \href{http://jmlr.org/proceedings/papers/v28/wan13.html}{Regularization of
  Neural Networks using DropConnect}, 2013.
\newblock JMLR W\&CP 28 (3) : 1058--1066, 2013.

\bibitem{FastDropout}
Sida Wang and Christopher Manning.
\newblock
  \href{http://jmlr.csail.mit.edu/proceedings/papers/v28/wang13a.html}{Fast
  dropout training}.
\newblock {\em JMLR W \& CP}, 28(2):118--126, 2013.

\bibitem{RecuurentNetworkRegularization}
Wojciech Zaremba, Ilya Sutskever, and Oriol Vinyals.
\newblock Recurrent neural network regularization, 2014.
\newblock
  \href{http://arxiv.org/abs/1409.2329}{http://arxiv.org/abs/1409.2329}.

\end{thebibliography}

\clearpage{}

\appendix

\section{Fast dropout}

We might have called batchwise dropout \emph{fast dropout }but that
name is already taken\emph{ }\cite{FastDropout}. Fast dropout is
an alternative form of regularization that uses a probabilistic modeling
technique to imitate the effect of dropout; each hidden unit is replaced
with a Gaussian probability distribution. The \emph{fast }relates
to reducing the number of training epochs needed compared to regular
dropout (with reference to results in a preprint%
\footnote{\href{http://arxiv.org/abs/1207.0580}{http://arxiv.org/abs/1207.0580}%
} of \cite{dropout}). Training a network 784-800-800-10 on the MNIST
dataset with 20\% input dropout and 50\% hidden-layer dropout, fast
dropout converges to a test error of 1.29\% after 100 epochs of L-BFGS.
This appears to be substantially better than the test error obtained
in the preprint after 100 epochs of regular dropout training.

However, this is a dangerous comparison to make. The authors of \cite{dropout}
used a learning-rate scheme designed to produce optimal accuracy eventually,
\emph{not} after just one hundred epochs. We tried using batchwise dropout with minibatches of size 100 and an annealed
learning rate of $0.01e^{-0.01\times\mathrm{epoch}}$. We trained
a network with two hidden layers of 800 rectified linear units each. Training
for 100 epochs resulted in a test error of 1.22\% (s.d. 0.03\%). After
200 epochs the test error has reduced further to 1.12\% (s.d. 0.04\%).
Moreover, per epoch, batchwise-dropout is faster than regular dropout
while fast-dropout is slower. Assuming we can make comparisons across
different programs%
\footnote{Using our software to implement the network, each batchwise dropout
training epoch take 0.67 times as long as independent dropout. In
\cite{FastDropout} a figures of 1.5 is given for the ratio between
fast- and independent-dropout when using minibatch SGD; when using
L-BFGS to train fast-dropout networks the training time per epoch
will presumably be even more than 1.5 times longer, as L-BFGS use
line-searches requiring additional forward passes through the neural
network.}, the 200 epochs of batchwise dropout training take less time than the 100 epoch of fast dropout training.

\end{document}